\documentclass[letterpaper]{article} % DO NOT CHANGE THIS
\usepackage{aaai24}  % DO NOT CHANGE THIS
\usepackage{times}  % DO NOT CHANGE THIS
\usepackage{helvet}  % DO NOT CHANGE THIS
\usepackage{courier}  % DO NOT CHANGE THIS
\usepackage[hyphens]{url}  % DO NOT CHANGE THIS
\usepackage{graphicx} % DO NOT CHANGE THIS
\usepackage{natbib}  % DO NOT CHANGE THIS AND DO NOT ADD ANY OPTIONS TO IT
\usepackage{caption} % DO NOT CHANGE THIS AND DO NOT ADD ANY OPTIONS TO IT
\usepackage{times}  % DO NOT CHANGE THIS
\usepackage{helvet}  % DO NOT CHANGE THIS
\usepackage{courier}  % DO NOT CHANGE THIS
\usepackage[hyphens]{url}  % DO NOT CHANGE THIS
\usepackage{graphicx} % DO NOT CHANGE THIS
\usepackage{natbib}  % DO NOT CHANGE THIS AND DO NOT ADD ANY OPTIONS TO IT
\usepackage{caption} % DO NOT CHANGE THIS AND DO NOT ADD ANY OPTIONS TO IT
\usepackage{colortbl}
\usepackage{enumitem}
\DeclareCaptionStyle{ruled}{labelfont=normalfont,labelsep=colon,strut=off} % DO NOT CHANGE THIS
\usepackage{amsmath,amsfonts,amsthm,bm}
\usepackage{subfigure}
\usepackage{caption}
\usepackage{multirow}
\usepackage{bbding}
\usepackage{bibentry}
\usepackage{booktabs} 
\usepackage{newfloat}
\usepackage{listings}
\usepackage{pifont} % star
\usepackage{arydshln}
\usepackage{xcolor}

\usepackage{algorithm}
\usepackage{algorithmic}

\usepackage{newfloat}
\usepackage{listings}
\DeclareCaptionStyle{ruled}{labelfont=normalfont,labelsep=colon,strut=off} % DO NOT CHANGE THIS
\floatstyle{ruled}
\newfloat{listing}{tb}{lst}{}
\floatname{listing}{Listing}
\title{Earthfarseer: Versatile Spatio-Temporal Dynamical Systems Modeling \\ in One Model}

\author{
    Hao Wu\textsuperscript{\rm 1}\thanks{This work is completed during a research internship at HKUST (Guangzhou) (2022.03-2022.06).},
    Yuxuan Liang\textsuperscript{\rm 2},
    Wei Xiong\textsuperscript{\rm 3 \dag},
    Zhengyang Zhou\textsuperscript{\rm 1}, 
    Wei Huang\textsuperscript{\rm 4}, \\
    Shilong Wang\textsuperscript{\rm 1},
    Kun Wang\textsuperscript{\rm 1}\thanks{Corresponding author.}
}

 \affiliations{
    \textsuperscript{\rm 1}University of Science and Technology of China \\
    \textsuperscript{\rm 2}Hong Kong University of Science and Technology (Guangzhou) \\
    \textsuperscript{\rm 3}Tsinghua University
    \textsuperscript{\rm 4}University of Tokyo \;\;

    easylearninghao@gmail.com; yuxliang@outlook.com; xiongw21@mails.tsinghua.edu.cn; zzy0929@ustc.edu.cn; \\
    weihuang.uts@gmail.com; wslong2000@gmail.com; wk520529@mail.ustc.edu.cn
}

\usepackage{bibentry}
\begin{document}
\maketitle

\begin{abstract}
Efficiently modeling spatio-temporal (ST) physical processes and observations presents a challenging problem for the deep learning community. Many recent studies have concentrated on meticulously reconciling various advantages, leading to designed models that are neither simple nor practical. To address this issue, this paper presents a systematic study on existing shortcomings faced by off-the-shelf models, including \textit{lack of local fidelity}, \textit{poor prediction performance over long time-steps}, \textit{low scalability}, and \textit{inefficiency}. To systematically address the aforementioned problems, we propose an EarthFarseer, a concise framework that combines parallel local convolutions and global Fourier-based transformer architectures,  enabling  dynamically capture the local-global spatial interactions and dependencies. EarthFarseer also incorporates a multi-scale fully convolutional and Fourier architectures to efficiently and effectively capture the temporal  evolution.
Our proposal demonstrates strong adaptability across various tasks and datasets, with fast convergence and better local fidelity in long time-steps predictions. Extensive experiments and visualizations over eight human society physical and natural physical datasets demonstrates the state-of-the-art performance of EarthFarseer. We release our code at \url{https://github.com/easylearningscores/EarthFarseer}.
\end{abstract}

\section{Introduction}

Modeling spatio-temporal (ST) physical dynamics involves estimating states and physical parameters from a sequence of observations \cite{benacerraf1973mathematical, newell1980physical}. Generally, the understanding of a physical process is based on plenty of physical laws, such as Newton's second law \cite{pierson1993corpore} and Conservation of energy law \cite{sharan1996mathematical, egan1972numerical}. As tailor-made techniques, dynamical systems, primarily rooted in diverse physical systems, have been demonstrated to conform to most fundamental principles of real-world physical phenomena~\cite{chmiela2017machine, greydanus2019hamiltonian}, where such phenomena can be mostly recognized by existing mathematical frameworks.  To this end, the modeling of dynamic systems has increasingly become a generic approach that yields numerous versatile techniques for various applications. It is crucial in climate science for predicting environmental impacts, in economics and finance for market analysis, in engineering for system design, in public health for disease outbreak modeling, in neuroscience for understanding brain activity, in robotics for AI development, and in logistics for optimizing supply chains.~\cite{hale2012dynamics, humar2012dynamics}. Actually,  such dynamic systems can naturally model the time-varying evolution including both intricate natural system like meterology dynamics~\cite{wiggins2003introduction, humar2012dynamics, harish2016review}  and complex society system of  human mobility like traffic evolutions~\cite{ji2022stden, chen2022automated}.

\begin{figure}[!t]
\centering
\includegraphics[width=0.40\textwidth]{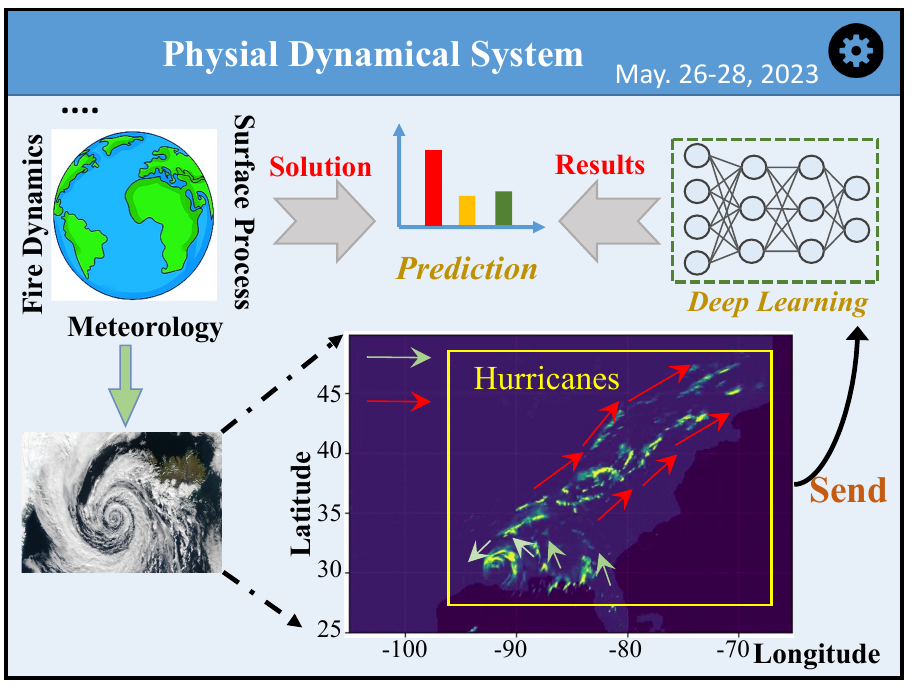}
\caption{A natural phenomenon in which global and local evolution are inconsistent. The hurricanes primarily exhibit clockwise rotation while in certain localized areas, the presence of convection results in the emergence of counterclockwise rotation.}
\label{fig:intro_1}
\end{figure}

\begin{figure*}[!h]
  \centering
  \includegraphics[width=0.95\linewidth]{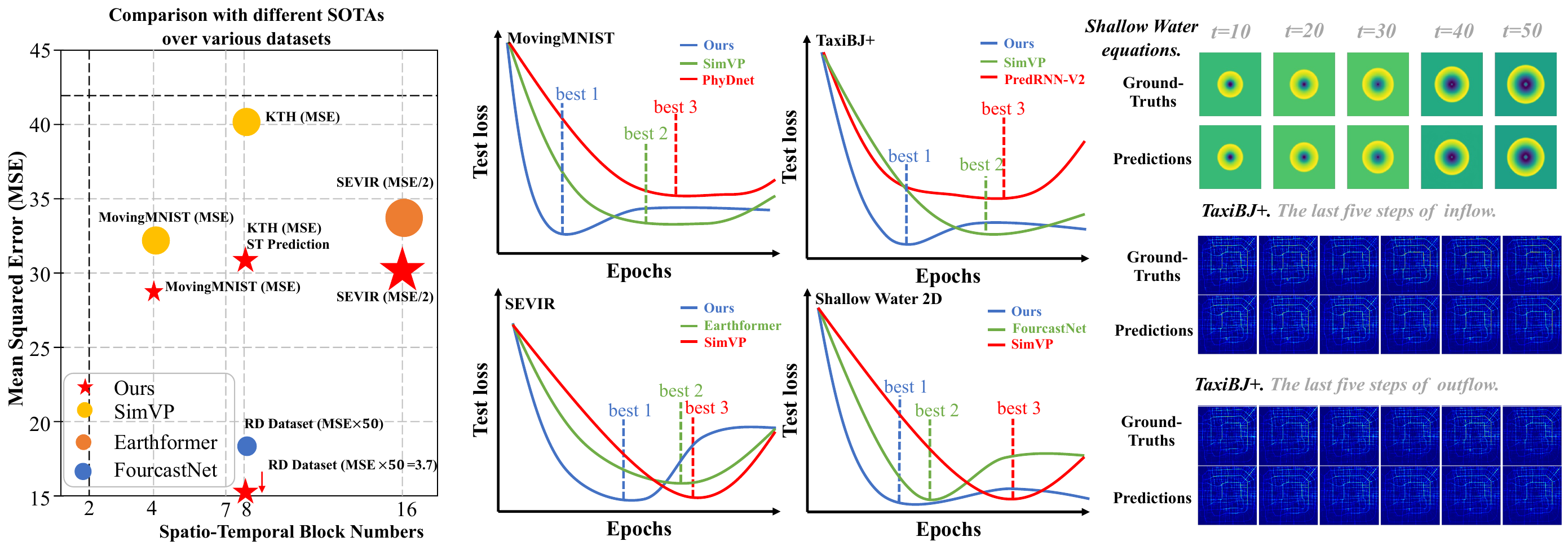}
  \caption{\textbf{\textit{Left.}} We showcase the performance comparisons between our model and SOTA models across diverse domains. \textbf{\textit{Middle.}} Convergence of our model compared to other models across different datasets. \textbf{\textit{Right.}} Our model demonstrates exceptional capability in addressing long-time steps prediction problems.}
  \label{fig:intro_2}
\end{figure*}

% In Shallow Water, we leverage 10 frames to accurately forecast the next 50 frames, here we provide visualizations showcasing the results for the entire ten-frame sequence. In TaxiBJ+, we adopt 12 frames to predict the next 12 frames, and we can find our model excels at accurately reproducing both inflow and outflow results.

Since dynamical systems are intrinsically tied to physical processes, they theoretically adhere to the constraints imposed by partial differential equations (PDEs). However, modeling and figuring out the above-mentioned dynamics with inherent physical theories is complicated and intractable to resolve. Fortunately, due to the similar properties of both spatial dependencies and temporal evolution between a general physical process and spatio-temporal modeling, we can opportunely attribute the physical process  to a spatio-temporal representation learning problem~\cite{yang2022learning, pathak2022fourcastnet}.

Intriguingly, the fascination of machine learning with physical phenomena has significantly intensified in recent years \cite{shi2015convolutional, greydanus2019hamiltonian}. Many research efforts have turned to video understanding \cite{chang2021mau, gao2022simvp} and physical-guided \cite{jia2021physics, lu2021learning} deep learning to capture ST characteristics in a data-driven manner. These approaches usually design various spatial or temporal components tailored for effectively characterizing a specific scenario or a dedicated task. Although many  ST frameworks deliver higher accuracy than the simple ones, they inevitably suffer from partial drawbacks ($\cal D$) outlined below:

    \noindent {${\cal D}1$}: \textbf{Lack of local fidelity.} Given the inconsistency and distinctive local-global dependencies in dynamic systems, existing modules usually focus on global regularity but fail to preserve the \textit{local fidelity} -- local dynamics may differ from global dynamics.
    %(See example in Fig~\ref{fig:intro_1} and Fig~\ref{fig:fire} in Appendix A). 

    \noindent {${\cal D}2$}: \textbf{Poor predictions over long time-steps.} Complex and continuous dynamic systems often exhibit intricate temporal correlations, leading to poor performance in long-term predictions \cite{isomura2021dimensionality}.

    \noindent {${\cal D}3$}: \textbf{Low scalability.} The intricate and convoluted component designs confine the model only capable to resolve specific tasks (e.g., super-resolution \cite{liang2019urbanfm}, flow prediction \cite{pan2019urban}), leading to limited scalability.

    \noindent {${\cal D}4$}: \textbf{Inefficiency.} The well-designed but cumbersome ST blocks not only contribute to the inefficient training \cite{wang2023snowflake, wang2022searching} process but also pose challenges to the model deployment.

\noindent Consequently, the resultant models were neither simple nor practical. Worse still, simultaneously overcoming the aforementioned problems provides an obvious obstacle for existing models. This paper carefully examines and explores the initial systematic study on the aforementioned questions. We introduce a ST framework called EarthFarseer, which is unfolded through the design a universal and Fourier-based ST disentanglement solver, which is different from  methods that utilize Fourier operators exclusively in either the spatial or temporal domain \cite{AFNO}. The model overview and contributions are outlined as below:

To highlight, our model exhibits strong adaptability to a wide range of tasks, as well as different datasets encompassing natural physical and social dynamical systems. We place significant emphasis on the fact that our model showcases consistent and reliable results across diverse datasets through simple size scaling, thereby highlighting the inherent scalability of our approach (left side of Fig \ref{fig:intro_2}). 

For spatial correlations, we employ a parallel local convolution architecture and a global Fourier-based transformer (FoTF) to extract both local and global information. Subsequently, we perform both down-sampling and up-sampling to facilitate effective global-local information interaction and enhance the local fidelity. In our implementation, the fast Fourier transform (FFT) is exploited to transform the patchified two-dimensional outputs from temporal to frequency domain. Each frequency corresponds to a set of tensor values in the spatial domain, so we can quickly perform global perception. This guarantees an efficient model convergence (evidence illustrated in the middle of Fig \ref{fig:intro_2}).

For temporal correlations, we design a temporal dynamic evolution module, TeDev, which effectively captures the continuous dynamic evolution within low-dimension space. In comparison to traditional modeling of discrete static frames  \cite{walker2021predicting, wang2022predrnn}, our model undergoes a transformation from the continuous time domain to the frequency domain through Fourier transformation, better preserving the long-term dependence of spatio-temporal data. Further, different from the prevailing neural ordinary differential equation (ODE) algorithms \cite{park2021vid}, our model showcases the great prominence in efficiency and long-term prediction tasks when captures intricated dynamics of systems without differential equation based nonlinear features. Through the exploitation of a low-parameter linear convolution projection, TeDev can effciently and accurately predict arbitrary future frames (right side of Fig \ref{fig:intro_2}).

\begin{figure*}[t]
  \centering
  \includegraphics[width=1\linewidth]{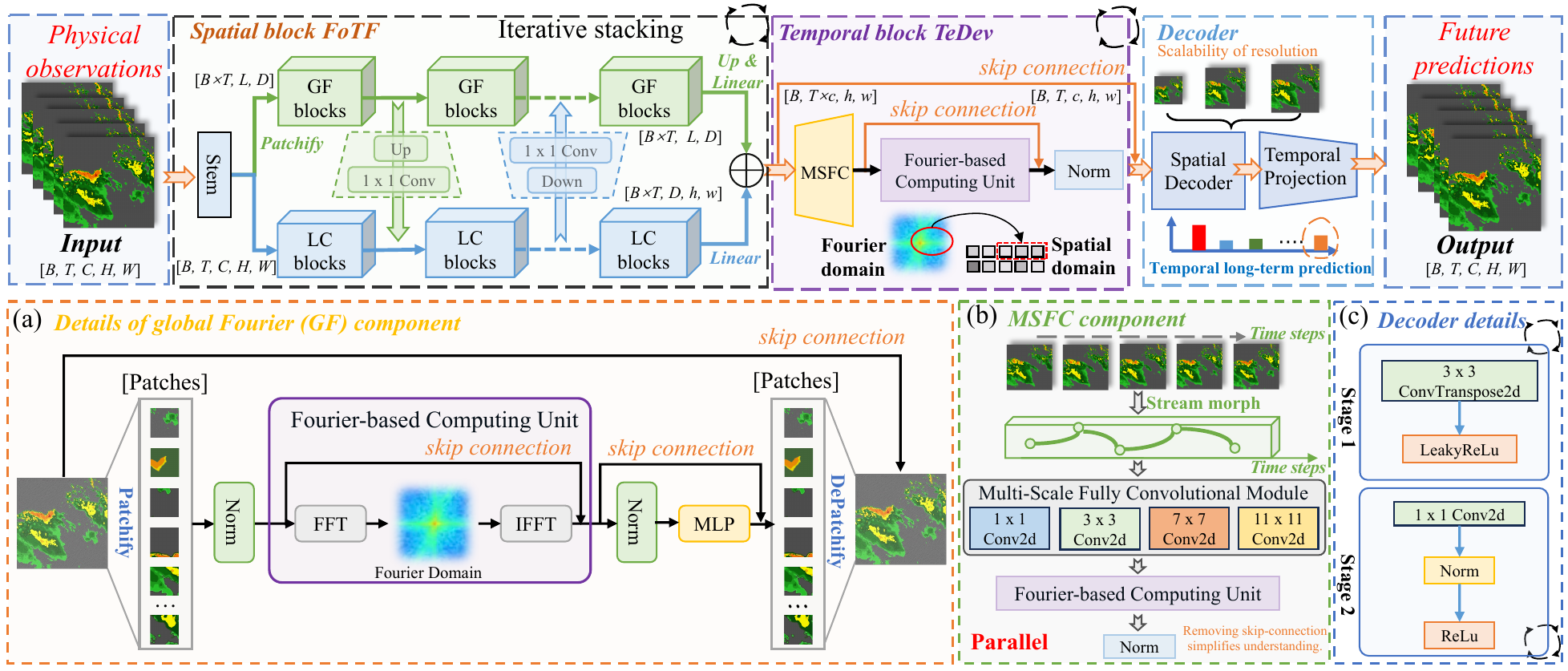}
  \caption{The upper half of the image presents an overview of the model, where Fig (a), (b), and (c) respectively showcase the details of the spatial module, temporal module, and decoding module.}
  \label{fig:mainmodel}
\end{figure*}

\section{Related Work}
% Our work shares common ground with several lines of research, and we will now summarize these studies below.

\subsection{Spatio-temporal prediction methods} can be roughly divided into CNN-based \cite{oh2015action, mathieu2015deep, tulyakov2018mocogan}, RNN-based \cite{wang2023brave, wang2022predrnn}, and other models including the combinations \cite{weissenborn2019scaling, kumar2019videoflow} and transformer based  models \cite{dosovitskiy2020image, bai2022rainformer}. While there are several existing models based on graph neural networks (GNNs), their primary focus is on handling graph data \cite{wang2020deep, wang2022a2djp}, which go out of the scope of our work.

\subsection{Video Prediction} has become a crucial research topic in the multimedia community, resulting in the proposal of numerous methods to tackle this challenge. Early studies primarily focused on analyzing spatio-temporal signals extracted from RGB frames \cite{shi2015convolutional}. Recently, there has been a growing interest in integrating video prediction with external information such as optical flow, semantic maps, and human posture data~\cite{liu2017video, pan2019video}. However, in real-world applications, accessing such external information may not always be feasible~\cite{wu2023pastnet, anonymous2023nuwadynamics}. Moreover, the current solutions still exhibit suboptimal efficiency and effectiveness when dealing with high-resolution videos. In this study, we concentrate on modeling continuous physical observations, which can sometimes be interpreted as a video prediction task.

\section{Methodology}

As depicted in Fig~\ref{fig:mainmodel}, EarthFarseer comprises three primary components: the FoTF spatial module, the TeDev temporal module, and a decoder. Going beyond ST components, we highlight decoder ability to handle spatial scale expansion and arbitrary length prediction in the temporal domain. we will present the preliminaries and elaborate on the contributions of each of our modules towards achieving local fidelity, model scalability, and SOTA performances.

\subsection{Preliminaries}
Dynamical systems describe how a system evolves from its current state to future states. Specifically, with an environment space $X$ and a state space $S \subset X$, they're modeled by a function $F$.
\begin{equation}  \label{system}
\small
\boldsymbol{X}_{t+\mathrm{d} t}=F\left(\boldsymbol{X}_t\right), \quad t=0,  dt, 2dt, 3dt, \ldots
\end{equation}

where $X_{t} \in S$ is the system's current state at time t, which can be further understood as a snapshot. Each snapshot contains $C$ color channels within a spatial resolution of $H \times W$. $dt$ represents the time increment. $F: S \rightarrow S$ describes the state evolution of the system from $X_{t}$ to its successive state $X_{t+dt}$. We consider the discrete time system, as any time-continuous system can be discretized with an appropriate $dt$.

\subsection{Spatial block FoTF}

FoTF contains two branches, namely, the local CNN module (LC block) and the global Fourier-based Transformer (GF), which work in parallel to enhance global perception and local fidelity. We model continuous physical observations as image inputs with a batch size of $B$ and a time length of $T$. Then, the input dimension is $[B, T, C, H, W]$ and we feed inputs into a Stem module, which includes a 1x1 successive convolution for initial information extraction.

\subsubsection{Local CNN Branch} utilizes a CNN architecture to capture fine-grained features. Due to the strong inductive bias of CNNs, we employ smaller convolutional kernels, \textit{i,e,,} 3x3, to enable the model with local perceptual capabilities. Specifically, LC is composed of $N_{e}$ ConvNormRelu unit:

\begin{equation}\small
\begin{gathered}
    {\rm{Z}}_{LC}^{i + 1} = {\rm{LeakyRelu}}( {{\rm{GNorm}}( {{\rm{Conv}2{\rm{d}}}( {{\rm{Z}}_{LC}^i} )} )})  \in \mathbb{R}^{[B \times T, D, h, w]} , \\ 1 \leq i \leq N_e 
\end{gathered}
\end{equation}

\noindent where $\rm{GNorm}$ and $\rm{LeakyRelu}$ denote group normalization and leaky relu function, respectively. ${\rm{Z}}_{LC}^{j + 1}$ denotes output from $j$-th LC block. Generally, LC block maps high-dimensional inputs into relatively low-dimensional representations ($H>h, W>w$), which will then be sent to temporal evolution module.

\subsubsection{Global Fourier-based Transformer} runs in parallel with the CNN branch, and in each stage, the size of the Transformer feature is consistent with the LC output feature. Let $X_{t} \in \mathbb{R}^{H \times W \times C}$ be an input observation at time $t$, the image is first tokenized into $L = HW/{p^2}$ non-overlapping patches with $p \times p$ patch size. Each patch is projected to an embedding $z \in \mathbb{R}^{D}$ by adopting a linear layer. Then we can obtain the tokenized image:
\begin{equation}\small
    Z_{t} =  ( z_{t}^{1};\ldots;z_{t}^{L} ) \in \mathbb{R}^{L \times D}
\end{equation}

\noindent where (; …; ) denotes row-wise stacking. For better understanding, we remove the subscript "t" to illustrate subsequent operations. To accommodate multiple batches and time steps inputs, we patchify them into dimensions, \textit{i.e.,} $[B \times T, L, D]$ compatible with the Transformer architecture as depicted above. Inspired by~\cite{guibas2021adaptive}, we replace multi-head self-attention (MSA) based token mixing with Fourier-based token mixing operator. Fourier transform converts an image from the spatial to the spectral domain, where each frequency corresponds to a set of spatial pixels. Therefore, the Fourier filter can process an image globally, rather than targeting a specific part like LC blocks. We conduct 2D real-valued fast Fourier transform (FFT) on patchified embedding ${\hat Z_{GF}}$:
\begin{equation} \small
    {\cal F} ({\hat Z_{GF}}) = \mathcal{F}[\hat Z_{GF}(x)] = \int_{-\infty}^{\infty} \hat Z_{GF}(x) e^{-2 \pi i k x} dx
    % {\hat Z_{GF}} \leftarrow {\cal F}\left( {{{\hat Z}_{GF}}} \right)
\end{equation}
where $k$ represents frequency, and $x$ represents the position within the spectral sequence, ${\hat Z_{GF}}$ is updated to its spectral domain representation with FFT. Similar to \cite{huang2023semicvt}, we take advantage of the conjugate symmetric property of the discrete Fourier transform and retain only half of the values for efficiency. After the Fourier transformation, we apply a linear transformation using a MLP. The purpose of this linear transformation is to map the frequency domain representation into a linear space, ensuring that the output has the same dimensionality as the input:
\begin{equation} \small
    {\cal F} ({\hat Z_{GF}}) \leftarrow {\rm MLP}({\cal F} ({\hat Z_{GF}}))
\end{equation}

We replace MSA with MLP, which can significantly reduce the computational complexity from $N^2 d+3 N d^2$ to quasi-linear $N d^2 / k+N d \log N$. Here $N$ refers to the sequence size (equal to the product of the height ($h$) and width ($w$) of the spatial grid.) and parameter $d$ represents the channel size. Here $m$ corresponds to the number of blocks used in GF component. The outputs from the MLP are then sent to an Inverse FFT (IFFT) module:

\begin{equation} \small
    {\hat Z_{GF}(x)} = {\mathcal{F}^{-1}}_k[F(k)](x) = \int_{-\infty}^{\infty} F^{+}(k) e^{2 \pi i k x} dk
\end{equation}

In this context, $\hat Z_{GF}(x)$ represents the output obtained after the IFFT, $F^{+}(k)$ denotes the result of the linear transformation. The primary objective of GF model is to transform the input in the spectral domain, map it to an output of the same dimensionality, and then revert it back to the time domain through the IFFT. This modul demonstrate the capability to effectively approximate global, long-range dependencies in higher resolution signals, all while avoiding the need for excessively deep architectures. We place model implement details in Appendix E.

\subsubsection{Global-local interactions.} We iteratively interact global GF and local LC modules multiple times to achieve information fusion. Concretely, we employ conv2d layer for upsampling (Up) and transposeconv2d for downsampling (Down), both with a kernel size of 3, stride of 1, and maintaining the same dimensions (The upper half of Fig \ref{fig:mainmodel}). In the final spatial block, we upsample the last GF output (with dimensions $[B \times T, L, D]$) and map it to the dimensions of $[B, T \times c, h, w]$ using a linear layer.

\subsection{Temporal block TeDev}
Fig \ref{fig:mainmodel}(b) illustrates the stream morph operator converting the discrete ST sequence into a continuous stream, processed through the MSFC module and Fourier-based unit for hidden feature extraction. This operator merges channel and time dimensions to transform discrete temporal dynamics into a continuous, irregular shape. The MSFC module leverages a multi-scale fully convolutional architecture with different kernel sizes for broad and detailed feature extraction—larger kernels capture global patterns, while smaller ones focus on local details. TeDev also includes an FFT/IFFT module (see Eq $4 \sim 6$) for transforming signals between time and frequency domains, enhancing signal analysis. TeDev effectively captures information across multiple time scales, ensuring temporal detail preservation in spatiotemporal predictions through its integrated approach.

\begin{equation}\scriptsize
\begin{gathered}
z^{j+1}=\sum_{k \in\{1,3,7,11\}} \operatorname{Conv2d}_{k \times k}\left(\operatorname{Conv2d}_{1 \times 1}\left(h^j\right)\right), \\
z^{j+1} = \mathcal{F}^{-1}{\rm(MLP(\mathcal{F}(z^{j+1})))} \in \mathbb{R}^{[B, T, c, h, w]} , N_e \leq j \leq N_e + N_t
\end{gathered}
\end{equation}
$N_t$ temporal blocks take the encoded hidden representation $z^j$ of the spatial encoder as input and obtain the hidden feature $z^{j+1}$ for the next time step. The feature is then processed by FFT/IFFT transform. In summary, TeDev's temporal evolution module comprehensively acquires features across scales from a continuously evolving time stack. It combines the output features of convolutional layers with diverse kernel sizes and performs spectral operations to ensure dimensional consistency.

\subsection{Decoder} Our decoder consists of two stages, \textit{i.e.,} spatial and temporal decoders, which allows for adaptation to different resolutions and flexible future time-step predictions.

% The first stage is the Spatial Decoder, which aims to project the hidden features back to the target dimension. The second stage is the Temporal Projection, which enables flexible prediction of frames at any time step. To avoid error accumulation that can arise from autoregressive prediction, we can easily change the length of the output prediction frames by expanding the temporal channel.

\subsubsection{Spatial Decoder} employs $N_d$ blocks to effectually reconstruct the latent features into an output of the desired size, which may assume any resolution. To be specific, it employs ConvTranspose2d for upsampling the encoded features to the target resolution, followed by the utilization of Tanh as the activation function to obtain the output. The layer combination form is explicated as follows:

% \begin{equation}
% \begin{array}{r}\small
% z^{d+1}=Tanh\left(\operatorname{ConvTranspose2d}\left(z^{d}\right)\right) \in \mathbb{R}^{[B, T, C, H, W]} , \\
% N_s+N_t<d \leq 2 N_s+N_t
% \end{array}
% \end{equation}

\begin{equation}\small
\begin{gathered}
z^{d+1}={\rm{Tanh}}(\operatorname{ConvTranspose2d}(z^{d})) \in \mathbb{R}^{[B, T, C, H, W]} , \\
N_e+N_t<d \leq N_e+N_t+N_d
\end{gathered}
\end{equation}

\subsubsection{Temporal Projection}
In order to flexibly predict future lengths, we utilize the ConvNormRelu unit to expand the time channel. Specifically, we concatenate the time and channel dimensions of the decoded features $z^{d+1}\in \mathbb{R}^{[B, T, C, H, W]}$ obtained in the first stage, resulting in a tensor of size $T \times C$, which is then mapped to $K \times C$, where $K$ is the desired target length, which can theoretically be any value. Subsequently, we perform a dimensional transformation on the resulting feature map to obtain the predicted target dimension $z^{d+2}\in \mathbb{R}^{[B, K, C, H, W]}$. The formal calculation process is as follows:

%通过扩展时间通道轻松地改变输出预测帧的长度

\begin{equation}\small
\begin{gathered}
    {z}^{d + 2} = {\rm{Relu}}( {{\rm{Norm}}( {{\rm{Conv}2{\rm{d}}}( {{z}^{d+1}} )} )}) \in \mathbb{R}^{[B, K, C, H, W]}
\end{gathered}
\end{equation}

Through the aforementioned spatio-temporal decoding module, we can output the results to specific resolutions and durations according to the requirements of specific prediction tasks, thus accommodating a wider range of needs.

\begin{table}[h]\scriptsize
\normalsize
  \centering
   \tabcolsep=0.4mm
    \begin{tabular}{l|ccccccc}
    \hline
    \textbf{Dataset} & \textbf{$N\_tr$} & \textbf{$N\_te$} & \textbf{($C, H, W$)} & \textbf{$I_l$} & \textbf{$O_l$} & \textbf{Interval} \\
    \hline
    MovingMNIST & 9000  & 1000  & (1, 64, 64) & 10    & 10 & -- \\
    TaxiBJ+ & 3555 & 445  & (2, 128, 128) & 12     & 12 & 30 mins\\
    KTH   & 108717 & 4086  & (1, 128, 128) & 10    & 20 & -- \\
    SEVIR & 4158  & 500   & (1, 384, 384) & 10    & 10 & 5 mins \\
    RainNet & 6000  & 1500   & (1, 208, 333) & 10    & 10 & 1 hour\\
    PD & 2000  & 500   & (3, 128, 128) & 6    & 6 & 5 seconds\\
    RD & 2000  & 500   & (3, 128, 128) & 2    & 2 & 1 second\\
    2DSWE & 4000  & 1000   & (1, 128, 128) & 50    & 50 & -- \\
    \hline
    \end{tabular}%
      \caption{Dataset statistics. $N\_tr$ and $N\_te$ denote the number of instances in the training and test sets. The lengths of the input and prediction sequences are $I_l$ and $O_l$, respectively.}
        \label{tab:dataset}
\end{table}

\section{Experiments}

\begin{table*}[!tb]
\Large
\setlength{\tabcolsep}{4pt}
    \resizebox{1\textwidth}{!}{
    \begin{tabular}{ccccccccccccccc}
    \toprule[1.5pt]
    \multirow{2}{*}{\large{Datasets}} &\multirow{2}{*}{\large{Metrics}} &\multicolumn{13}{c}{Models}      \\
    \cmidrule{3-15}
                &           & ConvLSTM & PredRNN-v2 & E3D-LSTM & SimVP & VIT    & SwinT & Rainformer & Earthformer & PhyDnet &  Vid-ODE & PDE-STD & FourcastNet & Ours\\
    \midrule\midrule
            & MSE    &103.3      &56.8            &41.3          &15.1  &62.1        &54.4     &85.8        &41.8     &24.4   &22.9     &23.1       &60.3         &\cellcolor{gray!30}{14.9}                  \\
MovingMNIST & MAE    &182.9      &126.1            &86.4          &49.8   &134.9       &111.7     &189.2       &92.8      &70.3   &69.2     &68.2      &129.8         &\cellcolor{gray!30}{33.2}                  \\
          
            \midrule
            & MAE    &5.5    &4.3       &4.1        &3.0         &3.4    & 3.2           &- -   &- -          &4.2    &3.9       &3.7        &- -   &\cellcolor{gray!30}{2.1}     \\
TaxiBJ+     & MAPE   &0.621    &0.469   &0.422      &0.307      &0.362    &0.306          &- -   &- -         &0.459    &0.413    &0.342         &- -            &\cellcolor{gray!30}{0.243}      \\
            \midrule
            & MSE    &126.2    &51.2       &86.2          &40.9       &57.4 & 52.1     &77.3      &48.2   &66.9         &49.8        &65.7        &102.1           &\cellcolor{gray!30}{31.8}      \\
KTH         & MAE    &128.3    &50.6       &85.6          &43.4       & 59.2    &55.3   &79.3     &52.3    &68.7        &50.1       &65.9       &104.9           &\cellcolor{gray!30}{32.9}    \\
            \midrule
            & MSE    &3.8         &3.9         &4.2	       &3.4        &4.4     &4.3     &4.0       &3.7        &4.8         &4.5       &4.4    &4.6         &\cellcolor{gray!30}{2.8}      \\
SEVIR       & CSI-M × 100 &41.9      &40.8    &40.4        &45.9       &37.1    &38.2    &36.6      &44.2       &39.4        &34.2      &36.2   &33.1     &\cellcolor{gray!30}{47.1}      \\
            \midrule
            & RMSE   &0.688  & 0.636 &0.613     &0.533    &0.472     &0.458   &0.533   &0.444   &0.533    &0.469         &0.463         &0.454         & \cellcolor{gray!30}{0.437}      \\
RainNet     & MSE    &0.472   &0.405  &0.376     &0.284    &0.223     &0.210   &0.284   &0.197   &0.282     &0.220        &0.215         &0.206         & \cellcolor{gray!30}{0.191}   \\           
            \midrule
            & MSE    &10.9  &9.6     &10.1    &5.4      &8.7    &8.4    &8.6        &7.2  &6.9       &4.8         &3.7         &5.1             &\cellcolor{gray!30}{2.2}      \\
PD          & MAE    &100.3 &95.4    &100.2     &50.9     &81.2   &79.5   &80.9       &73.4 & 68.7     &47.6         &38.9         &52.4             &\cellcolor{gray!30}{21.8}      \\            
            \midrule
            & MSE × 10    &21.2       &20.9         &18.2     &9.5   &13.2     &12.1     &9.7       &11.4        &10.8      &9.8         &9.8         &10.2             &\cellcolor{gray!30}{9.4}     \\
RD          & MAE         &52.7       &50.1         &42.6     &17.8  &27.3     &25.9     &43.2      &45.9        &22.6      &20.7        &20.3        &21.9             &\cellcolor{gray!30}{16.8}      \\          
            \midrule
            & MSE × 100 &11.2    & 8.9       &6.4       &3.1       &8.1     &7.6           &7.8       &7.4             &4.9         &4.5      &4.3      &5.2             &\cellcolor{gray!30}{2.6}      \\
2DSWE       & MAE       &54.3    & 53.1      &30.2      &17.2      &52.7    &50.3          &51.4      &49.2            &20.1        &19.8     &19.5     &21.7            &\cellcolor{gray!30}{10.5}      \\  \midrule           
\multicolumn{2}{c}{Avg Ranking}&6.2    & 3.5       &4.3       &3.3       &3.3     &3.5           &5.2       &4.7             &5.3         &4.6      &5.2      &4.8             &\cellcolor{gray!30}{1.7}      \\

    \bottomrule[1.5pt]              
    \end{tabular}}
    \caption{Model comparison with the state-of-the-arts over different evaluation metrics. We report the mean results from three runs. Given the distinct characteristics of various datasets, we present different dimensions across different rows to account for their unique properties.}
    \label{table:main1}
\end{table*}

In this section, we empirically demonstrate the superiority of our framework on seven datasets, including two human social dynamics system (TaxiBJ+, KTH \cite{schuldt2004recognizing}), five natural Scene dynamical systems (SEVIR~\cite{veillette2020sevir}, RainNet \cite{chen2022rainnet}, Pollutant-Diffusion (PD), Reaction-Diffusion) and 2D shallow water Equations (2DSWE)~\cite{takamoto2022pdebench}, and a synthetic systems (MovingMNIST \cite{srivastava2015unsupervised}). In the subsequent section, we will provide a detailed introduction to the dataset and baseline, along with the corresponding experimental settings and results.

\subsection{Experiment setting}

\subsubsection{Dataset Description} We conduct extensive experiments on eight datasets, including two human social dynamics system (II, III), five natural scene datasets (IV, V, VI, VII, VIII) and a synthetic datasets (I) in Tab \ref{tab:dataset}, for verifying the generalization ability and effectiveness of our algorithm. See dataset details in Appendix C and D.

% (I) \textbf{MovingMNIST} contains handwritten digits from the MNIST dataset. (II) \textbf{TaxiBJ+}: TaxiBJ contains trajectory data of Beijing from taxi GPS, with inflow and outflow channels. We extend TaxiBJ by collecting the latest trajectory from the resolution 32$\times$32 to 128$\times$128, named \textbf{TaxiBJ+}. (III) \textbf{KTH} includes 25 human performing six types of actions. By observing previous frames, the model can learn the dynamics of human motion and predict long-term posture changes in the future. (IV) \textbf{SEVIR} dataset contains radar-acquired measurements of vertical accumulation liquid (VIL), acquired every 5 minutes with a resolution of 1 km. (V) \textbf{RainNet} benchmark contains more than 62,400 pairs of high-quality low/high-resolution precipitation maps for over 17 years. (VI) \textbf{Pollutant-Diffusion (PD)} are obtained from the computational fluid dynamics (CFD) simulation results of pollutant dispersion. (VII) \textbf{Reaction-Diffusion (RD)} describes the ST evolution of material concentration in biological/chemical systems. Each image can be regarded as a equation solution. (VIII) \textbf{2D Shallow-Water Equations (2DSWE)} can be derived from the Navier-Stokes equation and used to model free surface flow problems. Each image can be regarded as a solution. 

\subsubsection{Baselines for Comparison} We compare EarthFarseer with the following baselines that belong to three categories:

    \noindent${{\cal B}_1}$. \textbf{Video Prediction Models}: We select ConvLSTM~\cite{shi2015convolutional}, PredRNN-v2~\cite{wang2022predrnn},  E3D-LSTM~\cite{Wang2019Eidetic3L} and SimVP-v2~\cite{tan2022simvp} as some of the most representative and advanced RNN architecture models in recent years.

    \noindent${{\cal B}_2}$. \textbf{Spatio-temporal Series Modeling}: We conduct experiments on advanced Transformer architecture models, including Vision Transformer~\cite{dosovitskiy2020image}, Swin Transformer~\cite{liu2021swin}, Rainformer~\cite{bai2022rainformer} and Earthformer~\cite{gao2022earthformer}. 
    
    % In particular, Earthformer is the most advanced model on the SEVIR dataset.

    \noindent${{\cal B}_3}$.\textbf{Physics-guided Neural Networks}: We use modeling methods that incorporate PDE or ODE in the model as baseline models for comparison, including Ad-fusion model, PhyDnet~\cite{guen2020disentangling}, Vid-ODE~\cite{park2021vid}, PDE-STD~\cite{dona2020pde}, and FourcastNet~\cite{pathak2022fourcastnet}, and FourcastNet represents a meteorological modeling model in the neural operator field.

\subsubsection{Evaluation Metrics} We train our model with mean squared error (MSE) metric. We further use MSE, mean absolute error (MAE), and mean absolute percentage error (MAPE) as common evaluation metrics. Additionally, for the SEVIR dataset, we add the CSI index \cite{ayzel2020rainnet} as a core metric for comparison. We place the metrics descriptions in Appendix B.

\subsubsection{Implementation Details} We implement our model using PyTorch framework and leverage the four A100-PCIE-40GB as computing support. In our paper, we generate \textbf{model configurations} (ours-$Ti/S/B$) by adjusting ST block numbers equal to 6,12,24. Concretely, tiny ($Ti$), small ($S$) and Base ($B$) models have $6$, $12$ and $24$ ST Blocks, respectively. Specifically, ST block can be divided into three sub-blocks, namely spatial encoder FoTF, temporal block TeDev and decoder block.

%We have placed the detailed descriptions in table \ref{tab:st} in Appendix C.

\subsection{Main Results}
In this subsection, we thoroughly investigate the scalability and effectiveness of EarthFarseer on various datasets. We conduct a comprehensive comparison of our proposal with video prediction, spatio-temporal series, and physical-guided models, for ST tasks on the social dynamics system, synthetic, and natural scene datasets (Tab \ref{table:main1}). We summarize our observations (obs) as follows:
\textbf{Obs 1.} \textbf{EarthFarseer consistently outperforms existing methods} under the same experimental settings over all datasets, verifying the superiority of our ST blocks via global-local modeling component and temporal Fourier design. \textbf{Obs 2.} Our model scales well to large datasets and performs well. For instance, on SEVIR dataset (64.83GB with resolution 384×384), we surpass the SOTA (Earthformer) 0.0287 on CSI-M metric, which \textbf{demonstrates the scalability of our proposal}. \textbf{Obs 3.} On certain physical datasets that PDE information, such as 2DSWE, physics-guided models like PDE-STD and FourcastNet outperform primarily video prediction models (except SimVP) and ST models, yielding superior results. Our model adeptly captures the underlying principles of PDEs, exhibiting a lower MAE index ranging from $0.5$ to $5.2$ compared to existing video prediction/PDE-based models.

\begin{figure*}[t]
  \centering
  \includegraphics[width=1\linewidth]{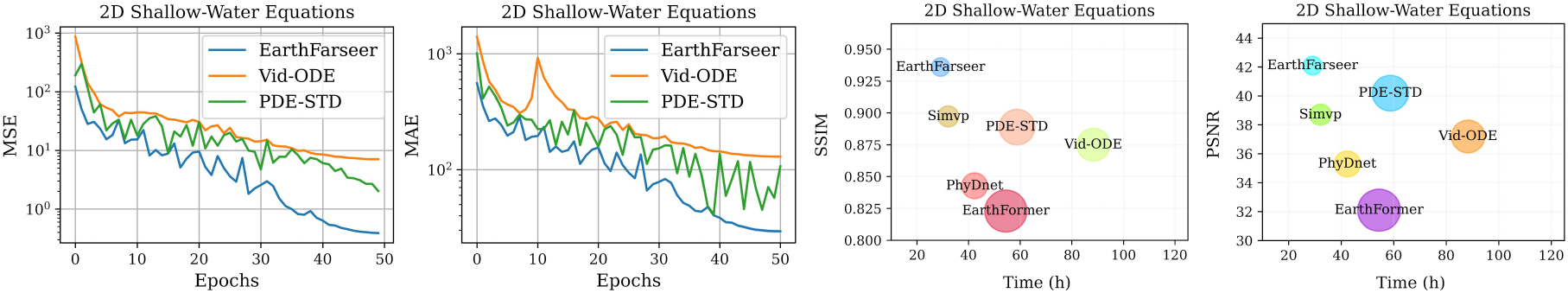}
  \caption{Model performance on 2DSWE dataset with different baselines. We measure the time it takes for the model to reach optimal performance by conducting fair executions across all frameworks on a Tesla V100-40GB.}
  \label{fig:scalab}
\end{figure*}

\subsection{Model Analysis}
\begin{figure}[h]
  \centering
  \includegraphics[width=0.85\linewidth]{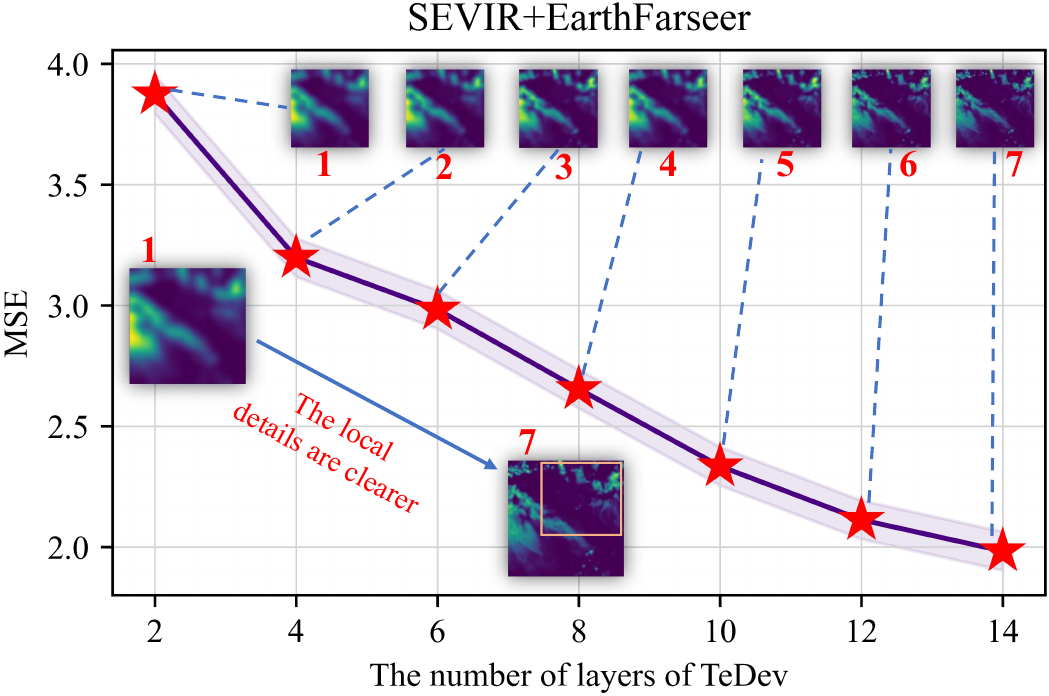}
  \caption{Model performance on SEVIR dataset with different number of temporal layers.}
  \label{fig:scalab}
\end{figure}

\subsubsection{${\cal Q}1$: Scalability analysis}
In our implementation, we can quickly expand the model size by stacking blocks layers. As shown in Tab \ref{table:main1}, we further explore the scalability of the temporal and spatial module. As meteorological exhibit highly nonlinear and chaotic characteristics, we selected the SEVIR to analyze the scalability of temporal component. We conduct experiments by selecting $2\sim14$ TeDev blocks layers, under settings with batch size as 16, training epochs as 300, and learning rate as 0.01 (Adam optimizer). The visualization results are presented quantitatively in the Fig~\ref{fig:scalab}. With the increase of the number of TeDev Blocks, we can observe a gradual decrease in the MSE index, and we can easily find that loacl details are becoming clearer as time module gradually increases. This phenomenon once again demonstrates the size scalability of our model. 

%Notably, our model incorporates a ViT-like architecture, offering intriguing possibilities for future expansion in a specific context.

We also conduct super-resolution experiments to further illustrate the scalability issues. In TaxiBJ+, we downsample the training data to $32\times32$ pixels to forecast future developments at $128\times128$ pixels over 12 steps. We discovered that the MAE was only 2.28 (MAPE=0.247). Compared to certain non-super-resolution prediction models, our results were significantly superior. This further demonstrates our model's capability in predicting spatial tasks across different resolutions (Fig 15\&16 for the visualization and training).

\subsubsection{${\cal Q}2$: Efficiency analysis} Due to the inherent complexity of solving PDEs, we select the 2DSWE dataset with PDE property as the benchmark for validating the efficiency of our model. As shown in Fig \ref{fig:scalab}, we can list observation: \textbf{Obs 1.} EarthFarseer presents a lower training error during the whole training process. \textbf{Obs 2.} Our model can achieve better convergence in faster training time. Specifically, we can save nearly 3/4 of the training time compared to VideoODE, which further verify the efficiency of EarthFarseer.

\subsubsection{${\cal Q}3$: Predicting future frames with flexible lengths} EarthFarseer can address the issues of accumulated error and delay effects in RNN-based models for predicting frames of arbitrary length. With a two-stage design, our approach restores the feature map to the input dimension, effectively preserving spatial feature information. Additionally, it utilizes a linear projection layer to expand the time channel, allowing for convenient adjustment of the output frame length. Evaluation on TaxiBJ+, PD and 2DSWE datasets reveals that EarthFarseer exhibits remarkable quantitative performance in experiments involving 10 $\to$ 30, 10 $\to$ 60 and 50 $\to$ 50 frames. We also showcase the 20 $\to$ 80 frames results on 2DSWE with different backbones, EarthFarseer outperforms baseline models the large margins, highlighting EarthFarseer's exceptional flexibility and prediction accuracy. These findings position EarthFarseer as a promising method in ST domains.

% Methods based on RNN, especially when predicting frames of arbitrary length, often suffer from two issues: accumulated error and delay effects. To address these issues, EarthFarseer's decoder employs a two-stage design that aims to solve these problems at the core. Specifically, stage 1 takes the feature map processed by the TeDev module and restores it to the input dimension, while preserving spatial feature information as much as possible via residual connections. Stage 2 uses a linear projection layer to expand the time channel through the TransposeConv2d module, allowing easy modification of the length of the predicted output frames.

% We evaluate EarthFarseer with the SEVIR and PD datasets and show that the model can learn natural fluid dynamics by observing previous frames and predict future fluid states over the long term. We compare EarthFarseer to baseline models specialized in the PDE field, including PhyDnet, Earthformer, PDE-STD, and FourcastNet, all trained for 100 epochs, and evaluate them using MSE and MAE. The experimental results show that EarthFarseer achieves excellent qualitative and quantitative performance in experiments of 10-30 and 10-50 frames, while the performance of other models decreases.

% Our research demonstrates that EarthFarseer has high flexibility and good prediction accuracy, making it a promising method in spatiotemporal prediction fields, particularly for handling complex fluid state changes.

\begin{figure}[h]
  \centering
  \includegraphics[width=1\linewidth]{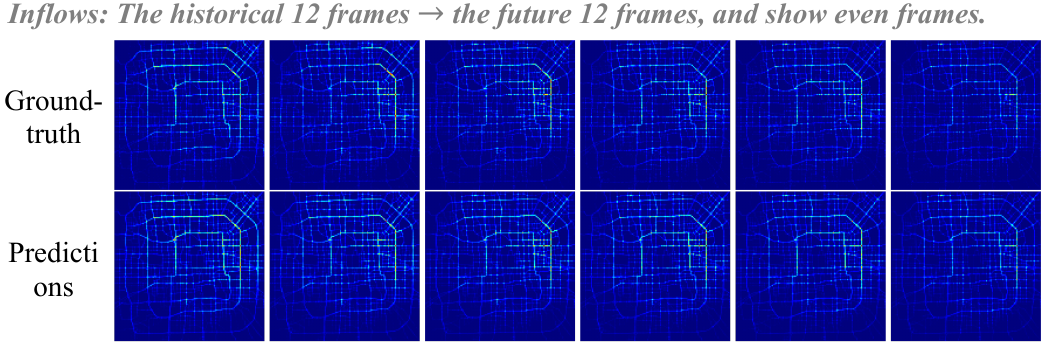}
  \label{fig:taxibj}
    \caption{Visualization of inflow and outflow prediction results on the TaxiBJ+ dataset.}
\end{figure}

\subsubsection{${\cal Q}4$: Local fidelity analysis}
\begin{figure}[h]
  \centering
  \includegraphics[width=0.95\linewidth]{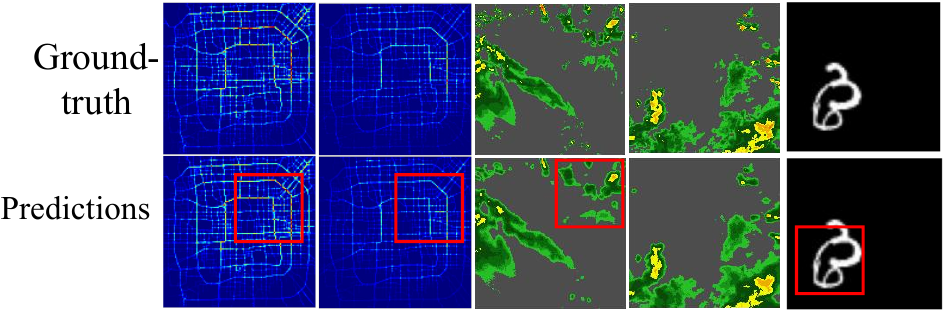}
  \caption{Visualizations of our framework on TaxiBJ+, SEVIR and MovingMNIST datasets, from which we can find that our model can preserve the local fidelity very well.}
  \label{fig:localmain}
\end{figure}

We proceed to consider another issue, \textit{i.e.,} local fidelity problem. We choose TaxiBJ+, SEVIR and MovingMNIST dataset as validation datasets. As shown in Fig \ref{fig:localmain}, our findings indicate that EarthFarseer effectively maintains local details while preserving the overall global context, particularly in the case of local outliers. This ability allows for the achievement of high fidelity in preserving local information. These results substantiate the local awareness exhibited by our model. 

%For ease of understanding, we present the more complete visualizations in Appendix D (Fig \ref{fig:localapp1} $\sim$ \ref{fig:compar}).

\subsection{Ablation Study}

In this part, we further explore the effectiveness of each individual component. In our settings, A denotes remove local convolutional component and D represents replace decoder with linear convolutional decoder. As shown in Tab \ref{tab:ablation}, our ablation experiments demonstrate that removing any module from our model leads to varying degrees of performance degradation. Both the local constraint (LC) and global constraint (FoTF) contribute to the model's performance on the spatial modules. For example, on the TaxiBJ+ dataset, removing the LC module leads to a decrease of 0.7 in the MAE metric, and removing the FoTF module leads to a decrease of 0.3. Our TeDev module outperforms the models using ViT and SwinT as a replacement for TeDev on the MovingMNIST and PD datasets. This suggests that our TeDev module is more suitable for tasks involving temporal information than using Transformer models.

\begin{table}[h] \small
\centering
\setlength{\tabcolsep}{3.2pt}
\begin{tabular}{l|c|c|c}
\hline
Method & MovingMNIST & TaxiBJ+  & RD \\ 
\hline
(A) Ours w/o LC & 19.7 & 2.8 & 14.1 \\ 
(B) Ours w/o FoTF & 16.6 & 2.4 & 17.2 \\ 
(C) Ours w/o TeDev & 22.1 & 3.1 & 17.8 \\ 
(D) Ours w/o Decoder & 15.9 & 2.2 & 10.2 \\ 
(E) Ours TeDev $\to$ ViT & 23.5 & 3.5 & 16.2 \\ 
(F) Ours TeDev $\to$ SwinT & 21.3 & 3.2 & 14.3 \\ 
\cline{1-4}
Ours (Full model) & \textbf{14.9} & \textbf{2.1} & \textbf{9.4} \\ 
\hline
\end{tabular}
\caption{Results of ablation experiments for different model structures on MovingMNIST, TaxiBJ+, and PD datasets. model effects were evaluated using MSE metrics for MovingMNIST and PD datasets and MAE metrics for TaxiBJ+ dataset.}
\label{tab:ablation}
\end{table}

\section{Conclusion}
This paper addresses the limitations of existing models that arise from the meticulous reconciliation of various advantages. We conducted a systematic study on the shortcomings faced by such models, including low scalability, inefficiency, poor long output predictions, and lack of local fidelity.  We propose a scalable framework that combines spatial local-global information extraction module and temporal dynamic evolution module. EarthFarseer demonstrates strong adaptability across various tasks and datasets, exhibiting fast convergence and high local fidelity in long-distance prediction tasks. Through extensive experiments and visualizations conducted on eight physical datasets, we showcase the SOTA performance of our proposal. All in all, our Earthfarseer achieves excellent performance.

\clearpage
\section{Acknowledgements}
This work is supported by Guangzhou Municiple Science and Technology Project 2023A03J0011. 

\bibliography{aaai24}

\end{document}